\documentclass{article}

\usepackage[final]{neurips_2019}

\usepackage[utf8]{inputenc}
\usepackage[T1]{fontenc}
\usepackage{natbib}
\usepackage{hyperref}
\usepackage{url}
\usepackage{booktabs}
\usepackage{amssymb}
\usepackage{amsfonts}
\usepackage{nicefrac}
\usepackage{microtype}
\usepackage{graphicx}
\usepackage{xcolor}
\usepackage{lipsum}
\usepackage{wrapfig}
\usepackage{xspace}
\usepackage{adjustbox}
\usepackage{tablefootnote}
\usepackage{caption}
\usepackage{subcaption}
\usepackage{tcolorbox}

\hypersetup{
    colorlinks=true,
    }

\title{
  Figuring out Figures: Using Textual References to Caption Scientific Figures \\
  \vspace{1em}
}

\author{
  Stanley Cao \\
  Department of Computer Science \\
  Stanford University \\
  \texttt{stanley.l.cao@stanford.edu} \\
  \And
  Kevin Liu \\
  Department of Computer Science \\
  Stanford University \\
  \texttt{liuk@stanford.edu} \\
}

\newcommand{\msc}{\textsc{MetaSciCap}\xspace}
\newcommand{\scicap}{\textsc{SciCap}\xspace}

\begin{document}

\maketitle

\begin{abstract}
  Figures are essential channels for densely communicating complex ideas in scientific papers. Previous work in automatically generating figure captions has been largely unsuccessful and has defaulted to using single-layer LSTMs, which no longer achieve state-of-the-art performance. In our work, we use the \scicap datasets curated by Hsu et al. \citep{scicap} and use a variant of a CLIP+GPT-2 encoder-decoder model with cross-attention to generate captions conditioned on the image. Furthermore, we augment our training pipeline by creating a new dataset \msc that incorporates textual metadata from the original paper relevant to the figure, such as the title, abstract, and in-text references. We use SciBERT to encode the textual metadata and use this encoding alongside the figure embedding. In our experimentation with different models, we found that the CLIP+GPT-2 model performs better when it receives all textual metadata from the SciBERT encoder in addition to the figure, but employing a SciBERT+GPT2 model that uses only the textual metadata achieved optimal performance.
\end{abstract}


\section{Introduction}

Image captioning has received significant attention from the computer vision and natural language processing communities. However, most research effort is directed toward captioning natural images \citep{DBLP:journals/corr/abs-2107-13114}. The captioning of computer-generated figures provides a very different challenge: it requires precise and often numerical data extraction, it uses different features (losing textures and object identities often used in region detection for captioning \citep{DBLP:journals/corr/abs-2107-13114}), and it requires generation in the specific genre of scientific writing. As described in Section \ref{sec:related}, few works have tackled this task in depth, with the leading work \citep{scicap} achieving a BLEU score hovering around 2.

Figure captioning also has useful applications: it may improve paper accessibility for the visually impaired; help authors write meaningful and high quality captions; and it may be a useful component to help text-based language models extract meaning from figures. More broadly, as language models begin to tackle more and more complex tasks that require deep domain understanding (e.g., Math Olympiad \citep{DBLP:journals/corr/abs-2202-01344}, competitive programming questions \citep{alphacode, hendrycksapps2021}), scientific figure captioning may be viewed as a proxy for the deeper goal of understanding academic research.

To tackle this task, we conjecture that looking at the figure alone provides insufficient information to write the caption. Intuitively, a caption is designed to \emph{augment} the figure and provide contextual information, unlike image captions (which are often similar to assistive descriptions). In light of this, we supplement our input with textual information, including paper metadata and in-text references. Using this approach, we see that text references improve performance significantly, although work remains to be done to ensure that image features are also efficiently incorporated.

\section{Related Work}
\label{sec:related}
Image captioning has received significant attention from the research community. However, although cutting-edge models have performed well on image captioning, these models have largely focused on captioning natural images such as the MS-COCO dataset \citep{DBLP:journals/corr/ChenFLVGDZ15}. As a result, they often use techniques like region detection through a Faster R-CNN \citep{7485869} to target interesting object regions for descriptions (e.g., OSCAR \citep{li2020oscar}, VIVO \citep{DBLP:journals/corr/abs-2009-13682}), a technique inappropriate for non-natural imagery. Notably, the Caption Transformer \citep{DBLP:journals/corr/abs-2101-10804} avoids region proposal networks and instead uses a fully transformer-based encoder-decoder architecture from pixels to caption; this approach partially inspired our present work.

Some research has also tackled diagram question-answering. Kembhavi et al. \citep{DBLP:conf/eccv/KembhaviSKSHF16} represent diagrams as a parse graph and extract knowledge using a multi-stage pipeline from diagrams in scientific textbooks, while Kim et al. do the same using a unified network \citep{Kim2018DynamicGG}. However, these approaches generally rely again on object localization and relation detection that is specialized toward diagram imagery.

The state of the art performance for learning from figures remains poor \citep{scicap}. Gomez-Perez and Ortega \citep{10.1145/3360901.3364420} learn a correspondence between figures and their captions using vision and language subnetworks; however, their work does not tackle generation. Previous work has attempted to caption figures based on synthetically-generated captions, but this has been criticized for being dissimilar to captions found in real scientific articles \citep{scicap}. As one of the few groups working on figure captioning, Hsu et al. \citep{scicap} create the \scicap dataset, a large-scale dataset of arXiv paper figures (see Section \ref{sec:data} for details) and establish a few baseline models with their dataset. They use a convolutional neural network (CNN) combined with an LSTM architecture, using the pre-trained ResNet-101 as the CNN to encode images into a 2048-dimensional vector. This image encoding is then projected to fit into the LSTM decoder, which used hidden layers of size 512. The authors design three variations of this baseline model, evaluating the caption quality using BLEU-4. When considering all the baseline models, the BLEU-4 scores hovers around 2, showing that the current state of the performance needs severe improvement. The authors also find that models trained on a subset of the dataset containing only single-sentence captions performed the best compared to first-sentence captions and captions with less than 100 words. The authors state that this is most likely because the single-sentence caption dataset collection had the smallest vocabulary size.

However, the CNN + LSTM approach scores very poorly, suggesting that more modern architectures might be able to improve on it significantly. One of the limitations of their model is its relatively small size and the fact that the CNN was trained on natural images, which does not match the image distribution that would appear in scientific papers. The \scicap authors also only tested a few input combinations, so testing further ones (even if unsuccessful) would help to shed light on what information is truly predictive of captions.

In the following sections, we present how we create a more expressive end-to-end model, augmented with additional textual information, that may outperform current CNN + LSTM models.

\section{Approach}

At the core of our approach is the idea of combining image perception with language generation. We model figure captioning as a sequence-to-sequence problem, using an encoder-decoder architecture with CLIP-ViT/B-32 \citep{DBLP:journals/corr/abs-2103-00020} as the encoder and either DistilGPT-2 or GPT-2 \citep{radford2019language} as the decoder. We chose CLIP as it was trained on a diverse distribution of visual input, including potentially-synthetic web images, which we felt might provide a better ground for fine-tuning than a model trained wholly on natural images such as ImageNet. CLIP uses a Vision Transformer architecture, which processes images by breaking up a $3 \times 224 \times 224$ image tensor into $32 \times 32$ patches. These patches are linearly embedded into token embeddings that are passed to a standard Transformer encoder. For experiments involving textual features (e.g., title, abstract, and references), we concatenate CLIP's output embeddings with those of SciBERT, a BERT encoder trained on scientific text \citep{Beltagy2019SciBERT}. Text features are tokenized and passed into SciBERT separately from the image encoder.

GPT-2 is a decoder-only model, which we augment to add encoder-decoder cross attention to the final hidden states of the encoder output. We chose GPT-2 as it was trained on a diversity of web text; we also ran an experiment with BART, an autoencoder version of BERT with an autoregressive decoder \citep{lewis-etal-2020-bart}, but this led to notably worse performance so we decided to use GPT-2 instead. (For some training runs, memory constraints require us to use DistilGPT-2, a distilled version of GPT-2 with 82M parameters vs. 117M for GPT-2.) We train end-to-end, passing images and metadata to the encoder and using teacher-forcing on the output of the decoder.

To implement and link these models, we fork the Huggingface Transformers library \citep{wolf-etal-2020-transformers}, adding CLIP support to the Vision Encoder-Decoder architecture and adding modifications to support passing arbitrary metadata to the encoder. Moreover, we incorporate the SciBERT encoder in our encoding pipeline, which works with CLIP to produce a concatenated embedding that is then given to GPT-2 or DistilGPT-2. In later experiments (see Section \ref{sec:results}), to incentivize our model to learn more from the figures, we used dropout on the SciBERT encodings with a dropout probability of 0.7 so that the model will be less inclined to rely on the textual metadata alone.
\citep{Falcon_PyTorch_Lightning_2019}.

As a baseline, we use the results as reported by Hsu et al. \citep{scicap}, which used a ResNet-101 CNN + single-layer LSTM with global attention, as well as a baseline that uses the first sentence of the first reference (if provided, otherwise empty string) as the predicted caption.\footnote{We first used a baseline that predicted the entire first reference (approx. 200 characters), but we found this to give artificially-inflated BLEU scores since the generated caption was typically much longer than the reference. This was supported by notably low ROUGE scores, which (unlike BLEU) weight recall. Since our reference captions are all one sentence, we hence decided to cap the reference baseline at one sentence.}

\section{Experiments}

\subsection{Data}
\label{sec:data}

We use the \scicap dataset \citep{scicap}, available on \href{https://github.com/tingyaohsu/SciCap}{GitHub}. It is an 18GB large-scale figure caption dataset based on Computer Science arXiv papers published between 2010 and 2020. It contains 416,000 graph plots (the most ubiquitous figure type) extracted from over 290,000 research papers. We use the SciCap-No-Subfig-Img subset, which denotes all figures that do not contain subfigures within them, and predict the first sentence of the caption. For preprocessing, each image is resized to $224 \times 224$ and normalized by the mean and std. dev. from the CLIP training dataset. We wrote a custom dataloader and preprocessor for this dataset.

Furthermore, Hsu et al. \citep{scicap} suggest that incorporating the paper's full text in which the figure belonged might boost performance of the model. However, we argue that there might only be certain features within the paper's text that might be useful, so we associate each figure with targeted text data, including the title, abstract, and references.

To associate paper metadata (title and abstract) with each figure, we linked this dataset with a publicly available arXiv metadata dump \citep{noauthor_arxiv_nodate}. To associate in-text references, we adapted \texttt{arxiv-public-datasets} \citep{clement2019arxiv} to extract full text from each PDF. We then masked out the original caption using the Striped Smith-Waterman algorithm for local sequence alignment and used a regular expression to extract a window of 100-characters on each side around each figure mention.

We call this augmented dataset (\scicap, metadata, and in-text references) \msc. Given an input of (figure, title, abstract, references), the model is expected to predict the caption as output. Note that when textual features are passed into the model, due to the limited context length of SciBERT, we use the first 100 characters of the title, 150 of the abstract, and allocate the rest of the context window for references. Each feature is separated by \texttt{[SEP]} tokens.

\subsection{Evaluation Method}

To generate captions, we use top-$p$ sampling with $p = 0.9$. We calculate a case-insensitive BLEU score against the reference caption using the standardized SacreBLEU \citep{post-2018-call}. We report all BLEU scores on a scale of 0-100. We also calculate a ROUGE-L score, which has two components: (1) precision, defined as the length of the longest common subsequence (LCS) between a reference $R$ and generated $C$ divided by the number of unigrams in the generated; and (2) recall, the length of the LCS divided by the number of unigrams in the reference. We report the F1 score, a metric that balances precision and recall. Our motivation to use ROUGE was to take into account recall in our evaluation, as our primary goal was to generate a caption that effectively confers the information in the reference. All results are reported against the \msc test split.

In addition to automated metrics, we report a sampling of qualitative behavior from the test split. See Section \ref{sec:results} for further details.

\subsection{Experimental Details}

We trained our models on a workstation with an NVIDIA RTX 2080 Ti and an Azure NC6s v3 VM with an NVIDIA V100. Select early models were trained on the Google TPU Research Cloud, but after experiencing PyTorch incompatibilities with the TPU architecture, we switched to other compute environments. We trained for 15 epochs, taking approximately 12 hours per experiment, and used AdamW with a learning rate of $5\times 10^{-5}$ and a linearly declining schedule. We ran experiments with other learning rate schedules, e.g., a fixed learning rate as well as a one cycle learning rate policy, but performance did not exceed a linear scheduler.

Training configurations are shown in Table \ref{tab:results}. SciBERT refers to \texttt{scibert\_scivocab\_uncased}; GPT-2 refers to GPT-2-base (117M parameters), and DistilGPT-2 refers to a pretrained distilled version of GPT-2-base with 87M parameters.

We encapsulate each of our model architectures in a custom PyTorch Lightning class, allowing for orchestration among multiple GPUs and supporting multithreaded dataloading \citep{Falcon_PyTorch_Lightning_2019}.

\subsection{Results}
\label{sec:results}

\begin{figure}[h]
    \centering
    \includegraphics[width=0.4\textwidth]{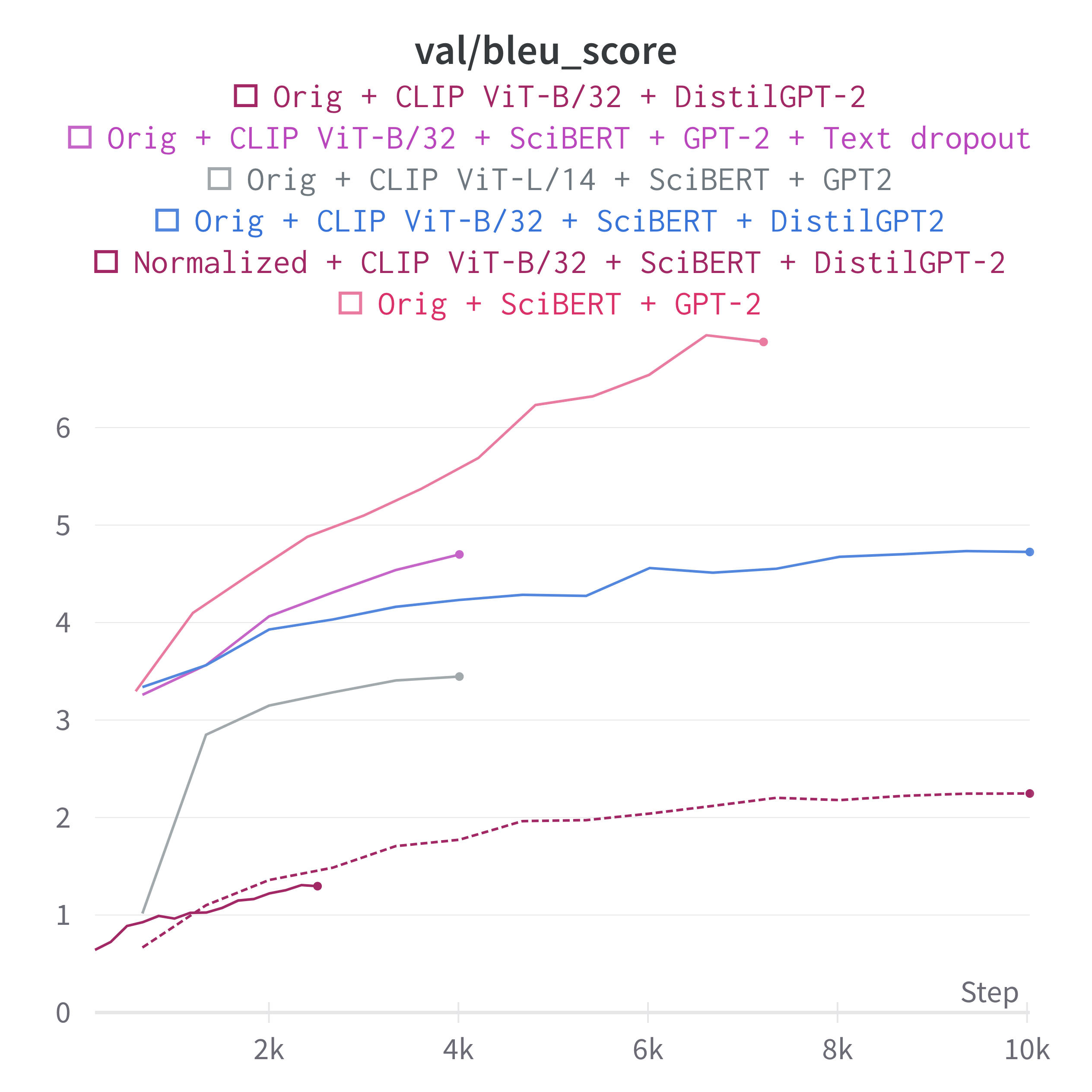}
    \includegraphics[width=0.4\textwidth]{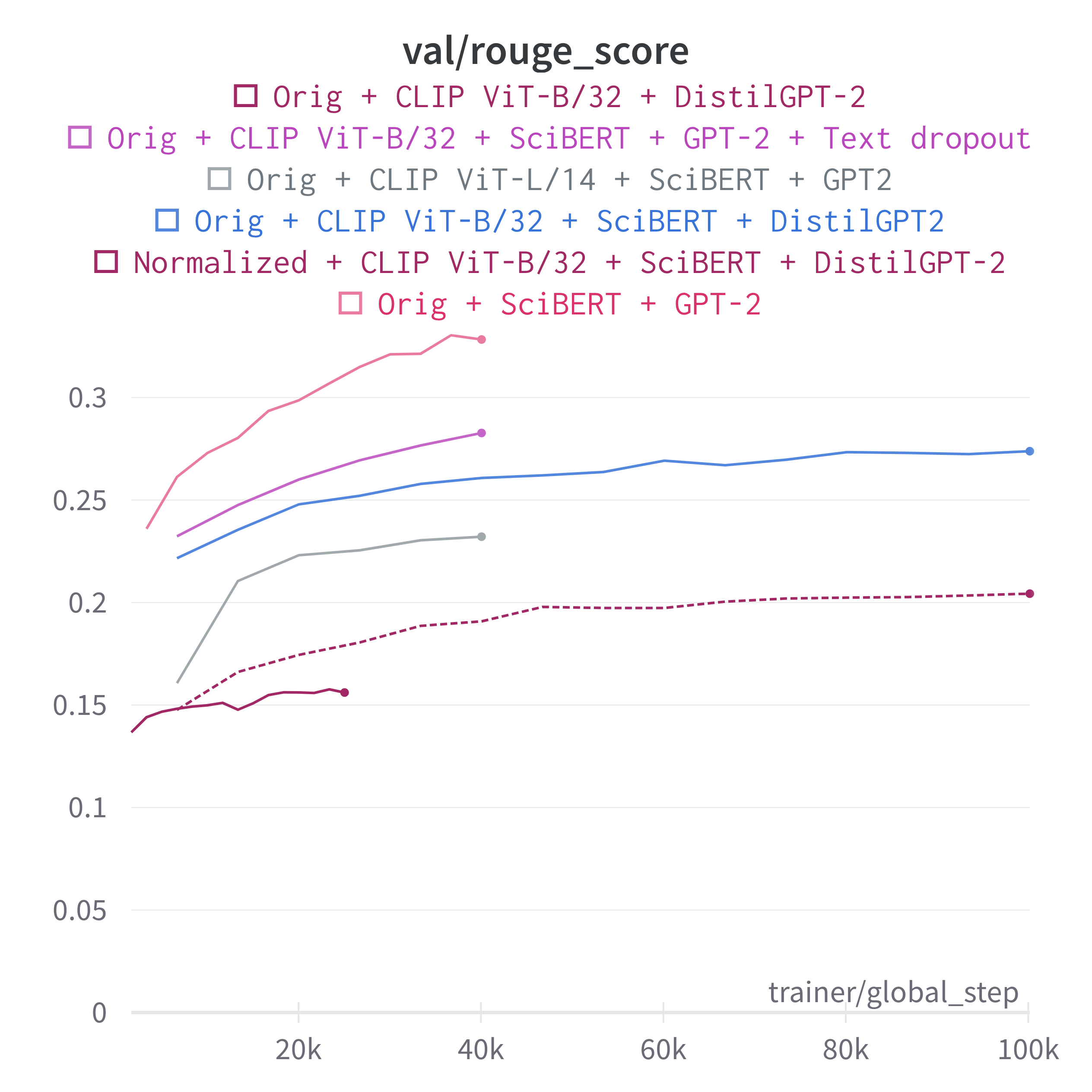}
    \caption{``Change in BLEU and ROUGE scores with the number of source predictions.''\\[1.5ex]
    \textit{\footnotesize This caption was actually autogenerated by our text-only model given our title, abstract, and reference. Best of 3 generations selected. In reality, this figure shows the improvement in BLEU and ROUGE over training.}}
    \label{fig:bleu_and_rouge}
\end{figure}

\begin{table}[h]
\centering
\begin{adjustbox}{center}
\small
\begin{tabular}{@{}llllll@{}}
\toprule
\textbf{Image?} & \textbf{Text?} & \textbf{Caption type} & \textbf{Model}                        & \textbf{BLEU (test) (0--100)} & \textbf{ROUGE-L F1 (test) (0--1)} \\ \midrule
\checkmark      &                & Normalized            & CNN + LSTM (SciCap)                   & 2.19                 & ---                        \\
                & \checkmark     & Normalized            & First reference sentence (baseline)   & 1.59                 & 0.09                       \\
\checkmark      & \checkmark     & Normalized            & CLIP ViT-B/32 + SciBERT + DistilGPT-2 &          \textbf{2.21}          &     \textbf{0.18}                     \\ \midrule
\checkmark      &                & Orig                  & \textit{CNN + LSTM (SciCap)}\tablefootnote{Hsu et al. did not report results on the First-Sentence captioning task with original captions and no-subfigures applied. As such, we have included here their results on First-Sentence, original captions, with \emph{subfigures included}, which is intended only for an intuitive sense of task difficulty and should not be seen as a definitive comparison of results. For a comparison of results, see our baselines.}                   & \textit{2.59}                 & --- \\
                & \checkmark     & Orig                  & First reference sentence (baseline)   & 1.38                     & 0.10                       \\
\checkmark      & \checkmark     & Orig                  & CLIP ViT-B/32 + SciBERT + DistilGPT-2 &         4.92             &           0.26                 \\
\checkmark      &                & Orig                  & CLIP ViT-B/32 + GPT-2                 &          1.02            &           0.13                 \\
                & \checkmark     & Orig                  & SciBERT + GPT-2                       & \textbf{6.71} &     \textbf{0.30}                       \\ \bottomrule
\end{tabular}
\end{adjustbox}
\caption{Results on the \msc test dataset, for both original and normalized captions. Results marked (SciCap) originate from the \scicap paper \citep{scicap}. For (SciCap) results, ROUGE-L F1 scores are not included as the original paper does not report ROUGE.}
\label{tab:results}
\end{table}


Our results are shown in Table \ref{tab:results}. Figure \ref{fig:bleu_and_rouge} also shows our BLEU and ROUGE scores on the validation set as a function of the training step.

On normalized captions, we achieve a BLEU score comparable to the CNN + LSTM baseline and BLEU and ROUGE scores that surpass our first-reference-sentence baseline. This is slightly worse than expected, likely due to normalized captions removing domain-specific knowledge as described in Section \ref{sec:analysis}.

However, for original captions, we see a significant improvement in performance over baselines. Our image + text model achieves a BLEU of 4.92 and ROUGE-L of 0.36, better than published SciCap results, our baseline, and unpublished results from Hsu et al. on a GPT-2 architecture \citep{kenneth_huang__survival_game_in-person_moment_2021}. Furthermore, a text-only ablation (not using CLIP-ViT-B/32 or an image input) taking in solely title, abstract, and references achieves a BLEU of 6.71 and ROUGE-L of 0.30. This suggests our model likely biases toward text features (and that image features seem to actually hurt performance), which we discuss in depth in Section \ref{sec:analysis}. We took two approaches to improve this imbalance: (1) trying a CLIP ViT-L/14 encoder, under the assumption that a larger encoder would be able to generalize better to figures (which did not improve performance, obtaining a BLEU score of 3.45, ROUGE 0.23); and (2) applying a strong dropout ($p = 0.7$) to the final outputs of the SciBERT encoder before running the GPT-2 decoder. This appeared to improve performance, but we were unable to obtain conclusive results due to compute and training time limitations. Nevertheless, this represents a potentially strong path forward.

\section{Analysis}
\label{sec:analysis}

Our key hypothesis is that reference data, representing the most relevant full-text excerpts for a given paper, should boost performance compared to the baselines described in Hsu et al \citep{scicap}. Our experimentation has shown that using references and a Transformer architecture leads to comparable results on normalized captions, but significantly better results on original captions (see Table \ref{tab:results}). We will analyze this phenomenon in Section \ref{sec:normvs}. In addition, we will discuss the possible reasons as to why the purely textual-based SciBERT+GPT-2 encoder-decoder model performs better than the CLIP+SciBERT+GPT-2 model.

\subsection{Normalized Captions vs. Original Captions}
\label{sec:normvs}
There are a few reasons why our model performs only comparably with the CNN + LSTM model on normalized captions, even when our model was supplied textual metadata. The text normalization process used in Hsu et al. mainly consisted of 2 strategies. The first approach was basic normalization, which essentially replaced all instances of numbers with \texttt{[NUM]} \citep{scicap}. The second approach was advanced normalization, which utilized regular expressions to replace any equations with \texttt{[EQUATION]}, and any text spans enclosed by any bracket pairs with \texttt{[BRACKET]} \citep{scicap}. This means many tokens that might have been included in the title, abstract, or figure references, were actually omitted from the normalized caption. Thus, our model could not effectively leverage the domain knowledge it may have learned from using the textual metadata, which means that our transformer architecture is essentially still captioning figures from the image embedding itself---the extra textual metadata proves to be of no use to our model. This supports the findings from Hsu et al.: the reason why the authors went with the CNN + LSTM model was because their experimentation with transformers did not result in any noticeable performance gain.

\begin{figure}[h]
    \centering
    \begin{subfigure}[t]{0.3\textwidth}
        \label{normalized:a}
        \vskip 0pt
        \centering
        \includegraphics[width=\textwidth]{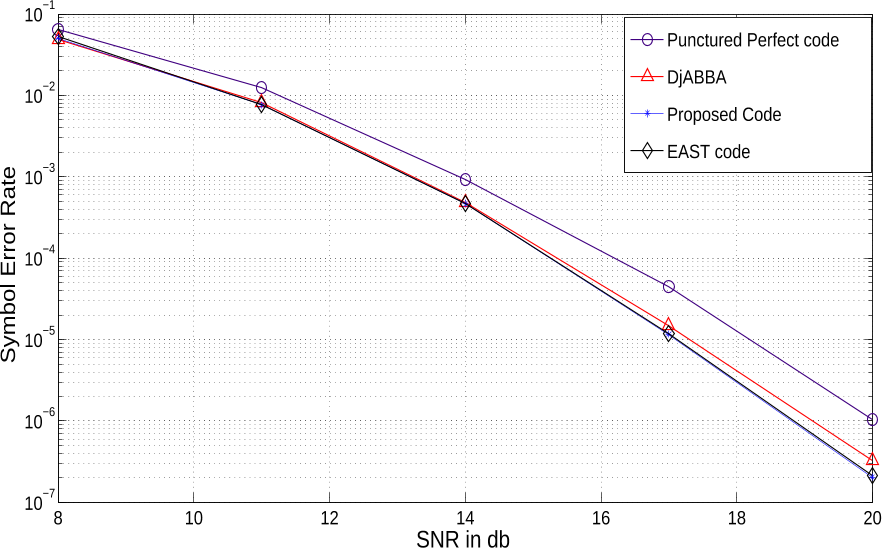}
        \caption{\textit{Gold Caption:} \\ ser performance at NUM bpcu for codes for4× systems. \\[1ex]
\textit{Predicted:} \\ ser versus snr at the ap for NUM NUM, NUM, and NUM NUM mimo.}
    \end{subfigure}
    \hfill
    \begin{subfigure}[t]{0.3\textwidth}
        \label{normalized:b}
        \vskip 0pt
        \centering
        \includegraphics[width=\textwidth]{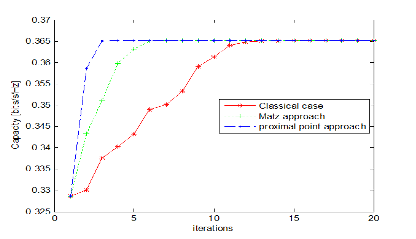}
        \caption{\textit{Gold Caption:} \\  comparision between the NUM approaches in the case of a dbsc channel.\\[1ex]
\textit{Predicted:} \\  lower bound on the rate : p BRACKET-TK = NUM - x/2}
    \end{subfigure}
    \hfill
    \begin{subfigure}[t]{0.3\textwidth}
        \label{normalized:c}
        \vskip 0pt
        \centering
        \includegraphics[width=0.7\textwidth]{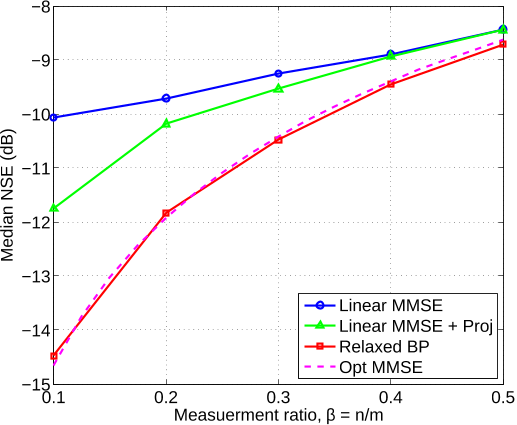}
        \caption{\textit{Gold Caption:} \\  relaxed bp algorithm with a gaussian prior and bounded noise output channel. the plot compares the simulated relaxed bp performance against [...] \\[1ex]
\textit{Predicted:} \\  median squared error over NUM trials for relaxed bp vs EQUAT-TK EQUAT}
    \end{subfigure}
    \caption{Model Predictions on Normalized Captions.}
    \label{fig:normalized}
\end{figure}

\begin{figure}[h]
    \centering
    \begin{subfigure}[t]{0.3\textwidth}
        \vskip 0pt
        \centering
        \includegraphics[width=\textwidth]{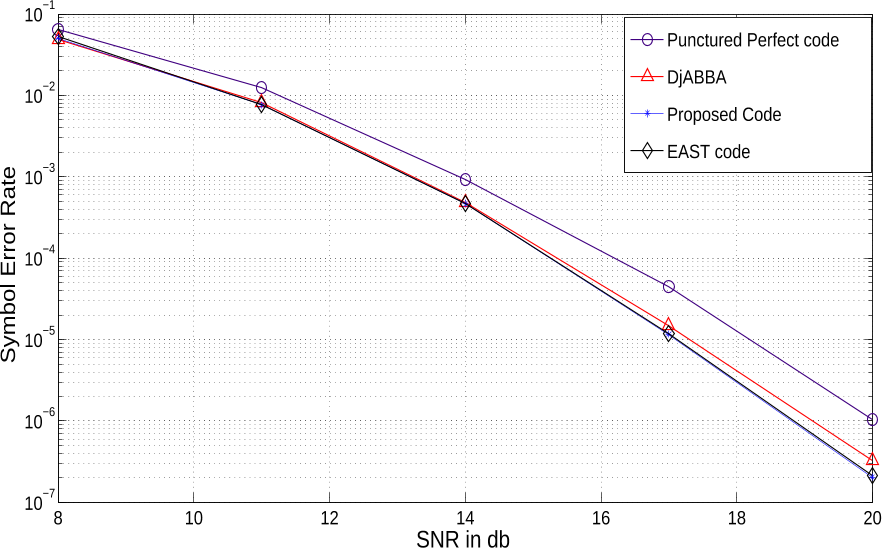}
        \caption{\textit{Gold Caption:} \\  Fig. 2. SER performance at 4 BPCU for codes for4× 2 systems \\[1ex]
\textit{Predicted:} \\  Fig. 2. SER performance by different schemes in 4× 2 MIMO Rayleigh fading}
        \label{orig:a}
    \end{subfigure}
    \hfill
    \begin{subfigure}[t]{0.3\textwidth}
        \label{orig:b}
        \vskip 0pt
        \centering
        \includegraphics[width=\textwidth]{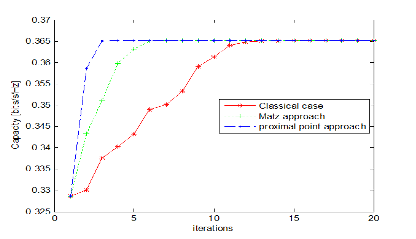}
        \caption{\textit{Gold Caption:} \\  Fig. 2. Comparision between the 3 approaches in the case of a DBSC channel\\[1ex]
\textit{Predicted:} \\  Fig. 9: Capacity results obtained in Algorithm 2 as a function of the number of users}
    \end{subfigure}
    \hfill
    \begin{subfigure}[t]{0.3\textwidth}
        \label{orig:c}
        \vskip 0pt
        \centering
        \includegraphics[width=\textwidth]{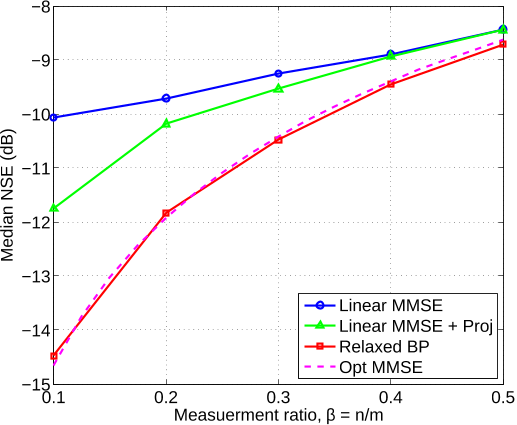}
        \caption{\textit{Gold Caption:} \\  Fig. 7. Relaxed BP algorithm with a Gaussian prior and bounded noise output channel. The plot compares the simulated relaxed BP performance against [...] \\[1ex]
\textit{Predicted:} \\  Fig. 7. SDP with bounded-norm approximation vs. sampling ratio in a Bernou}
    \end{subfigure}
    \caption{Model Predictions on Original Captions}
    \label{fig:orig}
\end{figure}

For a more detailed analysis of the internal workings, we provide a demonstration of our model on a few figures in the SCICAP test set. From Figure \ref{fig:normalized}, although our model learns of the bracket, equation, and number tokens (i.e., \texttt{BRACKET-TK}, \texttt{EQUAT-TK}, \texttt{NUM}), our model does not learn where to place them. This is most likely because these tokens do not appear in the textual metadata, so it is reasonable for our model to perform suboptimally on normalized captions, by virtue of it being trained on unnormalized metadata. For a more accurate assessment of our model architecture, normalization of the textual metadata would be required; this would result in training data that is suited for predicting \textit{normalized} captions, and we expect that our model will successfully specialize to this task.

Moreover, from Figure \ref{fig:normalized}, we observed that our model is inclined to predict equations, variables, and numbers, which might have appeared in the textual metadata. The normalized gold captions, however, do not have mathematical notation in the captions, so even though our model has acquired domain knowledge from the textual metadata, it is disincentivized from using it when generating the caption. This is supported by noting how in Figure \ref{orig:a}, the correct numerical expression $4 \times 2$ is predicted, demonstrating that our model would have used the domain knowledge (perhaps from the figure reference) when trained on original captions; supplemental trials (e.g., Figure \ref{fig:abbrev}) lead to a similar conclusion, as even exact acronyms used in the paper are replicated by our model's generated caption.\footnote{It is worthwhile to note that from the original captions in Figure \ref{fig:orig}, our model more or less predicts the correct figure number, contributing to a higher BLEU score.} Thus, the roughly similar performance of our model to the CNN + LSTM model on normalized captions is expected, since our model's ability to accurately predict elements of the original caption is not rewarded with a higher BLEU/ROUGE score. 

We are not as concerned with our model's performance on normalized captions for two main reasons. First, Hsu et al. found that there was no clear improvement in their CNN + LSTM model performance from training on normalized text, suggesting that textual normalization does not productively simplify the task for language models. Second, establishing a state of the art performance on original captions serves as a foundation for future work, because having the model predict normalized captions would omit equations and numbers that are important to understanding the figure---the goal of this task is to generate accurate, real captions rather than preprocessed ones.

\subsection{Performance Disparities from Modal Ablations}

\begin{figure}[h]
    \begin{tcolorbox}
        {\centering
        \includegraphics[width=0.4\textwidth]{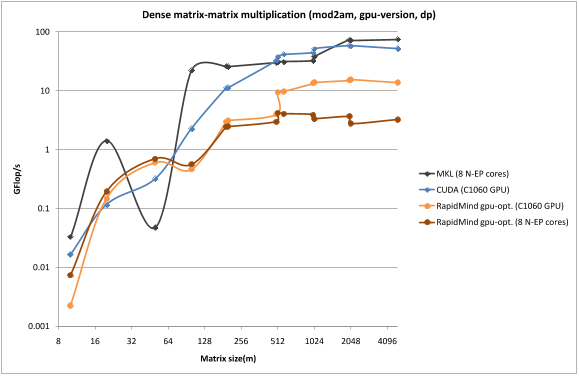}\\
        }
        \textit{References}: \\ s; a RapidMind version with improved performance of the cuda backend was scheduled for version 4.1. \textbf{Figure 3 shows the performance of the GPU-optimized version on various backends} and compares it with hardwar[SEP] [...] \\[1ex]
        \textit{Gold Caption:} \\ Fig. 3. \textbf{Performance comparision of the GPU-optimized version on various backends}. Performance measurements have been performed both on an Nvidia GPU and a Nehalem-EP socket [...] \\[1ex]
        \textit{Predicted:} \\ Fig. 3. \textbf{The performance comparison of the GPU-optimized version for a variety of back}
    \end{tcolorbox}
    \caption{
        Exemplary generation from text-only model showing the potential for memorization. If very similar wording is used in both the references and gold caption, the model may achieve strikingly good performance by simply regurgitating or slightly rewording reference text.
    }
    \label{fig:text-only-regurgitate}
\end{figure}

In general, as seen in Table \ref{tab:results}, adding textual metadata provided the greatest boost to performance, while the image-only models generally failed to surpass the baselines. The surprisingly strong performance of our model on the text modality thus merits investigation. One possible explanation for this is shown in Figure \ref{fig:text-only-regurgitate}, which shows a degenerate case where the caption can be derived from textual information alone. Essentially, while we masked out the original exact caption in our preprocessing, we took no action against text similar to the caption. Thus, if the reference text is very similar to the caption text (which may often occur in papers), the image is entirely unnecessary to caption the figure. It is arguable whether this represents a form of data poisoning or a legitimate textual feature; after all, in a use case such as an automatic caption generator, the model may very well already have references the author has written and so plagiarizing (if done well) would be acceptable.

Regarding the poor performance of CLIP+SciBERT+GPT-2 model compared to the SciBERT+GPT-2 model, one explanation may be a confounding variable. Due to compute limitations, for the CLIP+SciBERT+GPT-2 model, we had to use a DistilGPT-2 model as the decoder, which has been shown to have slightly lower performance than GPT-2-base. However, for the SciBERT+GPT-2 model, we were able to use the full GPT-2 model due to the omission of CLIP. This may have created a comparative gap in performance, supported by promising data on an initial training run with image-text and GPT-2 (scoring a BLEU of 4.54 in only 5/15 epochs); however, compute limitations again prevented us from completing this run. Nevertheless, this suggests that scaling up decoder size may be a simple way to improve BLEU performance.

The disparity in performance also suggests that the image data is being poorly utilized. One possible explanation for this may be the resizing and normalization performed in image preprocessing: figures in our dataset are often larger than $224 \times 224$, so shrinking them may hinder the model's ability to implicitly OCR text or detect lines. While we verified that the text remains readable to humans, this may not necessarily generalize to current models. It may also be that the difficulty of parsing a figure when starting from a natural image prior (as CLIP does for the most part) may be so great that the path of least resistance is to ignore the image, instead using figure metadata. If the image encoder is hence made ineffective, including it may only weaken the model by providing a noisy signal channel that the decoder must puzzle out. Further work could investigate alternative methods to encode the image that are better suited toward the vector nature of most figures, e.g., autovectorizing images and passing them in as SVG input to a text model.

\section{Conclusion}

Overall, we find that adding references as inputs to a figure captioning model has the potential to improve performance. We find that the transformer model architecture also achieves a better performance than the CNN + LSTM model when it is given textual metadata. However, because our model learns more from textual references, further experimentation should be done with more expressive image encoders, or perhaps improving the image encoder architecture. Pre-processing the image should also be investigated---e.g., vectorizing or extracting LaTeX/PostScript source for each image---because the vision encoder could leverage the patterns found from consistent inputs due to image pre-processing.

Furthermore, our experimentation remained limited to graph plots, while in reality, there are many different types of plots that models might need to caption. A more in-depth analysis on the difficulty of transfer learning for figure captioning across fields, time periods, and graph types could be an interesting task for further experiments. Other possible experiments include enlarging the decoder for more accurate captions, using a more expressive intermediate representation of figure images (e.g., graphs as done in \citep{DBLP:conf/eccv/KembhaviSKSHF16}), or better data processing to remove very similar reference text.

\newpage
\bibliographystyle{unsrt}
\bibliography{references}

\begin{thebibliography}{10}

\bibitem{scicap}
Ting-Yao Hsu, C~Lee Giles, and Ting-Hao Huang.
\newblock {S}ci{C}ap: Generating captions for scientific figures.
\newblock In {\em Findings of the Association for Computational Linguistics:
  EMNLP 2021}, pages 3258--3264, Punta Cana, Dominican Republic, November 2021.
  Association for Computational Linguistics.

\bibitem{DBLP:journals/corr/abs-2107-13114}
Ahmed Elhagry and Karima Kadaoui.
\newblock A thorough review on recent deep learning methodologies for image
  captioning.
\newblock {\em CoRR}, abs/2107.13114, 2021.

\bibitem{DBLP:journals/corr/abs-2202-01344}
Stanislas Polu, Jesse~Michael Han, Kunhao Zheng, Mantas Baksys, Igor
  Babuschkin, and Ilya Sutskever.
\newblock Formal mathematics statement curriculum learning.
\newblock {\em CoRR}, abs/2202.01344, 2022.

\bibitem{alphacode}
Yujia Li, David Choi, Junyoung Chung, Nate Kushman, Julian Schrittwieser, Rémi
  Leblond, Tom Eccles, James Keeling, Felix Gimeno, Agustin Dal~Lago, Thomas
  Hubert, Peter Choy, Cyprien de~Masson~d'Autume, Igor Babuschkin, Xinyun Chen,
  Po-Sen Huang, Johannes Welbl, Sven Gowal, Alexey Cherepanov, James Molloy,
  Daniel Mankowitz, Esme Sutherland~Robson, Pushmeet Kohli, Nando de~Freitas,
  Koray Kavukcuoglu, and Oriol Vinyals.
\newblock Competition-level code generation with alphacode, Feb 2022.

\bibitem{hendrycksapps2021}
Dan Hendrycks, Steven Basart, Saurav Kadavath, Mantas Mazeika, Akul Arora,
  Ethan Guo, Collin Burns, Samir Puranik, Horace He, Dawn Song, and Jacob
  Steinhardt.
\newblock Measuring coding challenge competence with apps.
\newblock {\em NeurIPS}, 2021.

\bibitem{DBLP:journals/corr/ChenFLVGDZ15}
Xinlei Chen, Hao Fang, Tsung{-}Yi Lin, Ramakrishna Vedantam, Saurabh Gupta,
  Piotr Doll{\'{a}}r, and C.~Lawrence Zitnick.
\newblock Microsoft {COCO} captions: Data collection and evaluation server.
\newblock {\em CoRR}, abs/1504.00325, 2015.

\bibitem{7485869}
Shaoqing Ren, Kaiming He, Ross Girshick, and Jian Sun.
\newblock Faster r-cnn: Towards real-time object detection with region proposal
  networks.
\newblock {\em IEEE Transactions on Pattern Analysis and Machine Intelligence},
  39(6):1137--1149, 2017.

\bibitem{li2020oscar}
Xiujun Li, Xi~Yin, Chunyuan Li, Xiaowei Hu, Pengchuan Zhang, Lei Zhang, Lijuan
  Wang, Houdong Hu, Li~Dong, Furu Wei, Yejin Choi, and Jianfeng Gao.
\newblock Oscar: Object-semantics aligned pre-training for vision-language
  tasks.
\newblock {\em ECCV 2020}, 2020.

\bibitem{DBLP:journals/corr/abs-2009-13682}
Xiaowei Hu, Xi~Yin, Kevin Lin, Lijuan Wang, Lei Zhang, Jianfeng Gao, and
  Zicheng Liu.
\newblock {VIVO:} surpassing human performance in novel object captioning with
  visual vocabulary pre-training.
\newblock {\em CoRR}, abs/2009.13682, 2020.

\bibitem{DBLP:journals/corr/abs-2101-10804}
Wei Liu, Sihan Chen, Longteng Guo, Xinxin Zhu, and Jing Liu.
\newblock {CPTR:} full transformer network for image captioning.
\newblock {\em CoRR}, abs/2101.10804, 2021.

\bibitem{DBLP:conf/eccv/KembhaviSKSHF16}
Aniruddha Kembhavi, Mike Salvato, Eric Kolve, Min~Joon Seo, Hannaneh
  Hajishirzi, and Ali Farhadi.
\newblock A diagram is worth a dozen images.
\newblock In {\em {ECCV} {(4)}}, volume 9908 of {\em Lecture Notes in Computer
  Science}, pages 235--251. Springer, 2016.

\bibitem{Kim2018DynamicGG}
Daesik Kim, Young~Joon Yoo, Jeesoo Kim, Sangkuk Lee, and Nojun Kwak.
\newblock Dynamic graph generation network: Generating relational knowledge
  from diagrams.
\newblock {\em 2018 IEEE/CVF Conference on Computer Vision and Pattern
  Recognition}, pages 4167--4175, 2018.

\bibitem{10.1145/3360901.3364420}
Jose~Manuel Gomez-Perez and Raul Ortega.
\newblock Look, read and enrich - learning from scientific figures and their
  captions.
\newblock In {\em Proceedings of the 10th International Conference on Knowledge
  Capture}, K-CAP '19, page 101–108, New York, NY, USA, 2019. Association for
  Computing Machinery.

\bibitem{DBLP:journals/corr/abs-2103-00020}
Alec Radford, Jong~Wook Kim, Chris Hallacy, Aditya Ramesh, Gabriel Goh,
  Sandhini Agarwal, Girish Sastry, Amanda Askell, Pamela Mishkin, Jack Clark,
  Gretchen Krueger, and Ilya Sutskever.
\newblock Learning transferable visual models from natural language
  supervision.
\newblock {\em CoRR}, abs/2103.00020, 2021.

\bibitem{radford2019language}
Alec Radford, Jeffrey Wu, Rewon Child, David Luan, Dario Amodei, Ilya
  Sutskever, et~al.
\newblock Language models are unsupervised multitask learners.
\newblock {\em OpenAI blog}, 1(8):9, 2019.

\bibitem{Beltagy2019SciBERT}
Iz~Beltagy, Kyle Lo, and Arman Cohan.
\newblock Scibert: Pretrained language model for scientific text.
\newblock In {\em EMNLP}, 2019.

\bibitem{lewis-etal-2020-bart}
Mike Lewis, Yinhan Liu, Naman Goyal, Marjan Ghazvininejad, Abdelrahman Mohamed,
  Omer Levy, Veselin Stoyanov, and Luke Zettlemoyer.
\newblock {BART}: Denoising sequence-to-sequence pre-training for natural
  language generation, translation, and comprehension.
\newblock In {\em Proceedings of the 58th Annual Meeting of the Association for
  Computational Linguistics}, pages 7871--7880, Online, July 2020. Association
  for Computational Linguistics.

\bibitem{wolf-etal-2020-transformers}
Thomas Wolf, Lysandre Debut, Victor Sanh, Julien Chaumond, Clement Delangue,
  Anthony Moi, Pierric Cistac, Tim Rault, Rémi Louf, Morgan Funtowicz, Joe
  Davison, Sam Shleifer, Patrick von Platen, Clara Ma, Yacine Jernite, Julien
  Plu, Canwen Xu, Teven~Le Scao, Sylvain Gugger, Mariama Drame, Quentin Lhoest,
  and Alexander~M. Rush.
\newblock Transformers: State-of-the-art natural language processing.
\newblock In {\em Proceedings of the 2020 Conference on Empirical Methods in
  Natural Language Processing: System Demonstrations}, pages 38--45, Online,
  October 2020. Association for Computational Linguistics.

\bibitem{Falcon_PyTorch_Lightning_2019}
William Falcon and {The PyTorch Lightning team}.
\newblock {PyTorch Lightning}, 3 2019.

\bibitem{noauthor_arxiv_nodate}
{arXiv} {Dataset}.
\newblock Kaggle.

\bibitem{clement2019arxiv}
Colin~B. Clement, Matthew Bierbaum, Kevin~P. O'Keeffe, and Alexander~A. Alemi.
\newblock On the use of arxiv as a dataset, 2019.

\bibitem{post-2018-call}
Matt Post.
\newblock A call for clarity in reporting {BLEU} scores.
\newblock In {\em Proceedings of the Third Conference on Machine Translation:
  Research Papers}, pages 186--191, Belgium, Brussels, October 2018.
  Association for Computational Linguistics.

\bibitem{kenneth_huang__survival_game_in-person_moment_2021}
{Kenneth Huang}.
\newblock The moment of truth: {If} you use a simple captioning model
  ({CNN}+{LSTM}), the resulting {BLEU} scores can be painfully low
  ({\textasciitilde}0.02). {We} tried {GPT}-2 (outside the paper), and the
  scores only slightly increased ({\textasciitilde}0.04). {So}: 1) {THIS} {IS}
  {A} {VERY} {HARD} {TASK} 2) {BLEU} is probably not a good metric
  https://t.co/{sNraHWREW2}, November 2021.

\end{thebibliography}
\newpage
\appendix

\section{Appendix}

\begin{figure}[h]
    \begin{tcolorbox}
        {\centering
        \includegraphics[width=0.4\textwidth]{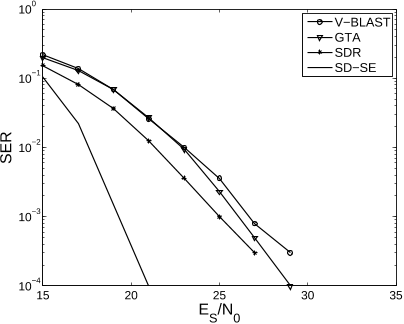}\\
        }
        \textit{Gold Caption:} \\ Fig. 5. Comparison of various detectors in12× 12 system, 16-QAM symbols. \\[1ex]
        \textit{Predicted:} \\ Fig. 5. MIMO detection performance for 16-QAM, 16-QAM
    \end{tcolorbox}
    \caption{
        Our model learns the exact acronyms used in figures.
    }
    \label{fig:abbrev}
\end{figure}

\end{document}